\documentclass{article}
\usepackage{spconf,amsmath,epsfig}
\usepackage{makecell}
\usepackage{multirow}
\let\OLDthebibliography\thebibliography
\renewcommand\thebibliography[1]{
  \OLDthebibliography{#1}
  \setlength{\parskip}{0pt}
  \setlength{\itemsep}{0pt plus 0.3ex}
}

\begin{document}\sloppy



\title{DefakeHop: A Light-Weight High-Performance Deepfake Detector \vspace{-0.1cm}}
%
\name{Hong-Shuo Chen$^{1}$, Mozhdeh Rouhsedaghat$^{1}$, Hamza Ghani$^{1}$, Shuowen Hu$^{2}$, Suya You$^{2}$, C.-C. Jay Kuo$^{1}$\vspace{-0.2cm}}
\address{University of Southern California, Los Angeles, California, USA$^{1}$ \\ Army Research Laboratory, Adelphi, Maryland, USA$^{2}$ \vspace{-0.3cm}}

\maketitle

\begin{abstract}

A light-weight high-performance Deepfake detection method, called
DefakeHop, is proposed in this work. State-of-the-art Deepfake
detection methods are built upon deep neural networks. 
DefakeHop extracts features automatically using the successive
subspace learning (SSL) principle from various parts of face 
images. The features are extracted by c/w Saab transform
and further processed by our feature distillation module 
using spatial dimension reduction and soft classification for each channel
to get a more concise description of the face.
Extensive experiments are conducted to demonstrate the effectiveness of
the proposed DefakeHop method. With a small model size of 42,845
parameters, DefakeHop achieves state-of-the-art performance
with the area under the ROC curve (AUC) of 100\%,
94.95\%, and 90.56\% on UADFV, Celeb-DF v1 and Celeb-DF
v2 datasets, respectively.

\end{abstract}
\section{Introduction}\label{sec:introduction}

As the number of Deepfake video contents grows rapidly, automatic
Deepfake detection has received a lot of attention in the
community of digital forensics. There were 7,964 Deepfake video clips
online at the beginning of 2019.  The number almost doubled to 14,678 in nine months. 
Deepfake videos can be potentially harmful to society, from non-consensual
explicit content creation to forged media by foreign adversaries used in
disinformation campaigns.  Besides the large quantity,
another major threat is that fake video quality has improved a lot over
a short period of time. The rise in quality makes it increasingly difficult for humans to detect the difference between real and fake videos without a side by side comparison. 
As a result, an automatic and effective Deepfake detection
mechanism is in urgent need. It is also desired to have a software
solution that runs easily on mobile devices to provide automatic warning
messages to people when fake videos are played. This is the objective of
our current research. 

Most state-of-the-art Deepfake detection methods are based upon deep learning (DL) technique and can be mainly categorized into two types: methods based on convolutional neural
networks (CNNs) \cite{tolosana2020deepfakes, rossler2019faceforensics++,
zhou2017two, afchar2018mesonet,nguyen2019multi,nguyen2019use} and
methods that integrate CNNs and recurrent neural networks (RNNs)
\cite{guera2018deepfake, sabir2019recurrent}.  While the former focuses
solely on images, the latter takes both spatial and temporal features
into account. DL-based solutions have several shortcomings. First, their size is typically large containing hundreds of thousands or
even millions of model parameters. Second, training them is computationally
expensive. There are also non-DL-based Deepfake detection methods \cite{yang2019exposing,
agarwal2019protecting, matern2019exploiting}, where handcrafted features
are extracted and fed into classifiers. The
performance of non-DL-based methods is usually inferior to that of DL-based
ones. 

A new non-DL-based solution to Deepfake detection, called
DefakeHop, is proposed in this work. DefakeHop consists of three main modules: 
1) PixelHop++, 2) feature distillation and 3) ensemble classification.
To derive the rich feature representation of faces, DefakeHop extracts features using PixelHop++ units \cite{chen2020pixelhop++} from various parts of face images.  The theory of PixelHop++ have been developed by Kuo {\em et al.} using SSL \cite{kuo2016understanding, kuo2019interpretable, chen2020pixelhop++}. PixelHop++ has been recently used for feature learning from low-resolution face images \cite{rouhsedaghat2020facehop, rouhsedaghat2020low} but, to the best of our knowledge,
this is the first time that it is used for feature learning from patches extracted from high-resolution color face images. Since features extracted by PixelHop++ are still not concise enough for classification, we also propose an effective feature distillation module to further reduce the feature dimension and derive a more concise description of the face. Our feature distillation module uses spatial dimension reduction to remove spatial correlation in a face and a soft classifier to include semantic meaning for each channel. Using this module the feature dimension is significantly reduced and only the most important information is kept. Finally, with the ensemble of different regions and frames, DefakeHop achieves state-of-the-art results on various benchmarks. 


The rest of this paper is organized as follows.  Background is
reviewed in Sec.  \ref{sec:review}. The proposed DefakeHop method is
presented in Sec.  \ref{sec:method}. Experiments are given in Sec. \ref{sec:experiments} to demonstrate the
effectiveness. Finally, conclusion is pointed out in Sec.  \ref{sec:conclusion}. 

\section{Background Review}\label{sec:review}

\subsection{Non-DL-based Methods}\label{subsec:non-DL}

Yang {\em et al.} \cite{yang2019exposing} exploited discrepancy between
head poses and facial landmarks for Deepfake detection. They first
extracted features such as rotational and translational differences
between real and fake videos and then applied the SVM classifier.
Agarwal {\em et al.} \cite{agarwal2019protecting} focused on detecting
Deepfake video of high-profile politicians and leveraged specific facial
patterns of individuals when they talk. They used the one-class SVM,
which is only trained on real videos of high profile individuals.
Matern {\em et al.} \cite{matern2019exploiting} used landmarks to find
visual artifacts in fake videos, e.g., missing reflections in eyes,
teeth replaced by a single white blob, etc. They adopted the logistic
regression and the multi-layer perceptron (MLP) classifiers. 

\subsection{DL-based Methods}\label{subsec:DL}

\quad {\bf CNN Solutions.} Li {\em et al.} \cite{li2019exposing} used CNNs to
detect the warping artifact that occurs when a source face is warped
into the target one. Several well known CNN architectures such as VGG16,
ResNet50, ResNet101, and ResNet152 were tried. VGG16 was trained from scratch while ResNet models were pretrained on the ImageNet dataset
and fine-tuned by image frames from Deepfake videos.  Afchar {\em et
al.} \cite{afchar2018mesonet} proposed a mesoscopic approach to Deepfake
detection and designed a small network that contains only 27,977
trainable parameters. Tolosana {\em et al.} \cite{tolosana2020deepfakes}
examined different facial regions and landmarks such as eyes, nose,
mouth, the whole face, and the face without eyes, nose, and mouth.
They applied the Xception network to each
region and classify whether it is fake or not.

{\bf Integrated CNN/RNN Solutions.} The integrated CNN/RNN solutions
exploit both spatial and temporal features. Sabir {\em et al.}
\cite{sabir2019recurrent} applied the DenseNet and the bidirectional RNN
to aligned faces. G{\"u}era and Delp \cite{guera2018deepfake} first
extracted features from a fully-connected-layer removed InceptionV3,
which was trained on the ImageNet. Then, they fed these features to an
LSTM (Long short-term memory) for sequence processing. Afterward,
they mapped the output of the LSTM, called the sequence descriptor, to a
shallow detection network to yield the probability of being a fake one.

\section{DefakeHop Method}\label{sec:method}

\begin{figure*}[h]
\centering
\includegraphics[width=1\textwidth]{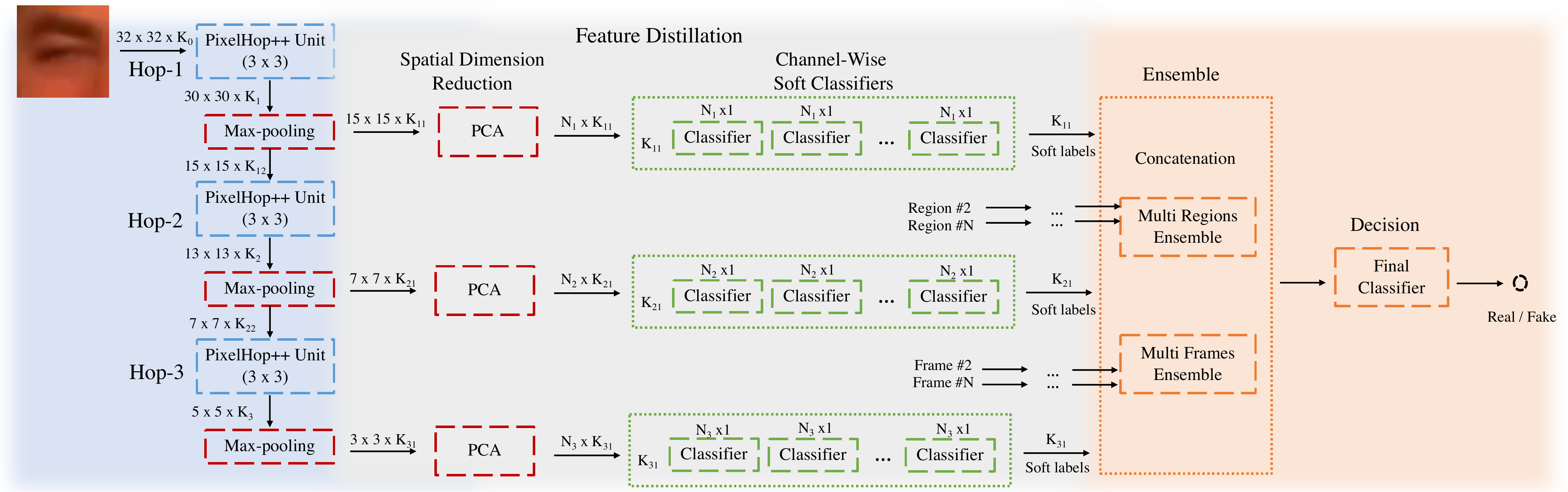}
\caption{An overview of the DefakeHop method.}\label{fig_ssl}
\end{figure*}

Besides the face image preprocessing step (see Fig. \ref{fig_pre}), the
proposed DefakeHop method consists of three main modules: 1) PixelHop++ 2) feature distillation and 3) ensemble classification. The
block diagram of the DefakeHop method is shown in Fig.  \ref{fig_ssl}.
Each of the components is elaborated below. 

\subsection{Face Image Preprocessing}\label{sec:preprocessing}

\begin{figure}[!t]
\centering
\includegraphics[width=\linewidth]{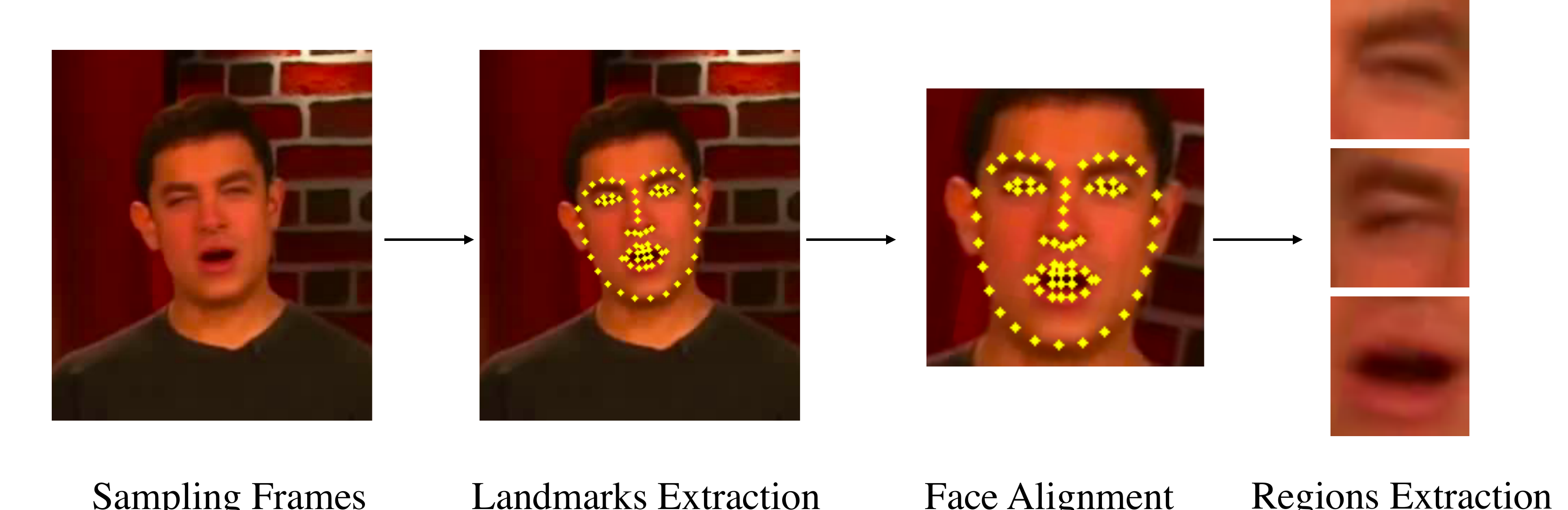}
\caption{Face image preprocessing.}\label{fig_pre}
\end{figure}

Face image preprocessing is the initial step in the DefakeHop system.
As shown in Fig. \ref{fig_pre}, we crop out face images from video
frames and then align and normalize the cropped-out faces to ensure proper and consistent inputs are fed to the
following modules in the pipeline. Preprocessing allows DefakeHop to
handle different resolutions, frame rates, and postures. Some details are
given below.  First, image frames are sampled from videos. Then,
68 facial landmarks are extracted from each frame by using an
open-source toolbox ``OpenFace2" \cite{baltrusaitis2018openface}.
After extracting facial landmarks from a frame, faces
are resized to $128 \times 128$ and rotated to specific coordinates to
make all samples consistent without different head poses or face sizes.
Finally, patches of size $32\times 32$ are cropped from different parts
of the face (e.g., the left eye, right eye and mouth) as the
input data to PixelHop++ module.

\subsection{PixelHop++ Module}\label{subsec:overview} 

PixelHop++ extracts rich and discriminant features from local blocks.
As shown in Fig. \ref{fig_ssl}, the input is a color
image of spatial resolution $32 \times 32$ that focuses on a particular
part of a human face. The block size and stride are hyperparameters for users to decide.
This process can be conducted in multiple stages to 
get a larger receptive field. The proposed DefakeHop system has three
PixelHop++ units in cascade, each of which has a block size of
$3\times3$ with the stride equal to one without padding.
The block of a pixel of the first hop contains $3 \times 3 \times
K_0= 9 K_0$ variables as a flattened vector, where $K_0=3$ for
the RGB input.

{\bf Spectral filtering and dimension reduction.} We exploit
the statistical correlations between pixel-based neighborhoods and apply
the c/w Saab transform to the flattened vector of dimension $K_1 = 9 K_0=27$ 
to obtain a feature representation of dimension $K_{11}$ (see discussion in
Sec. \ref{subsubsec:cwSaab}).  

{\bf Spatial max-pooling.} Since the blocks of two adjacent
pixels overlap with each other, there exists spatial redundancy between
them. We conduct the (2x2)-to-(1x1) maximum pooling unit to reduce the
spatial resolution of the output furthermore. 

\subsubsection{Channel-wise (c/w) Saab Transform}\label{subsubsec:cwSaab}

The Saab (subspace approximation via adjusted bias) transform
\cite{kuo2019interpretable} is a variant of PCA. It first decomposes a
signal space into the local mean and the frequency components and 
then applies the PCA to the frequency components 
to derive kernels. Each kernel represents a certain
frequency-selective filter. A kernel of a larger eigenvalue extracts a
lower frequency component while a kernel of a smaller eigenvalue
extracts a higher frequency component. The high-frequency components
with very small eigenvalues can be discarded for dimension reduction. 

\begin{figure}[!t]
\centering
\includegraphics[width=\linewidth]{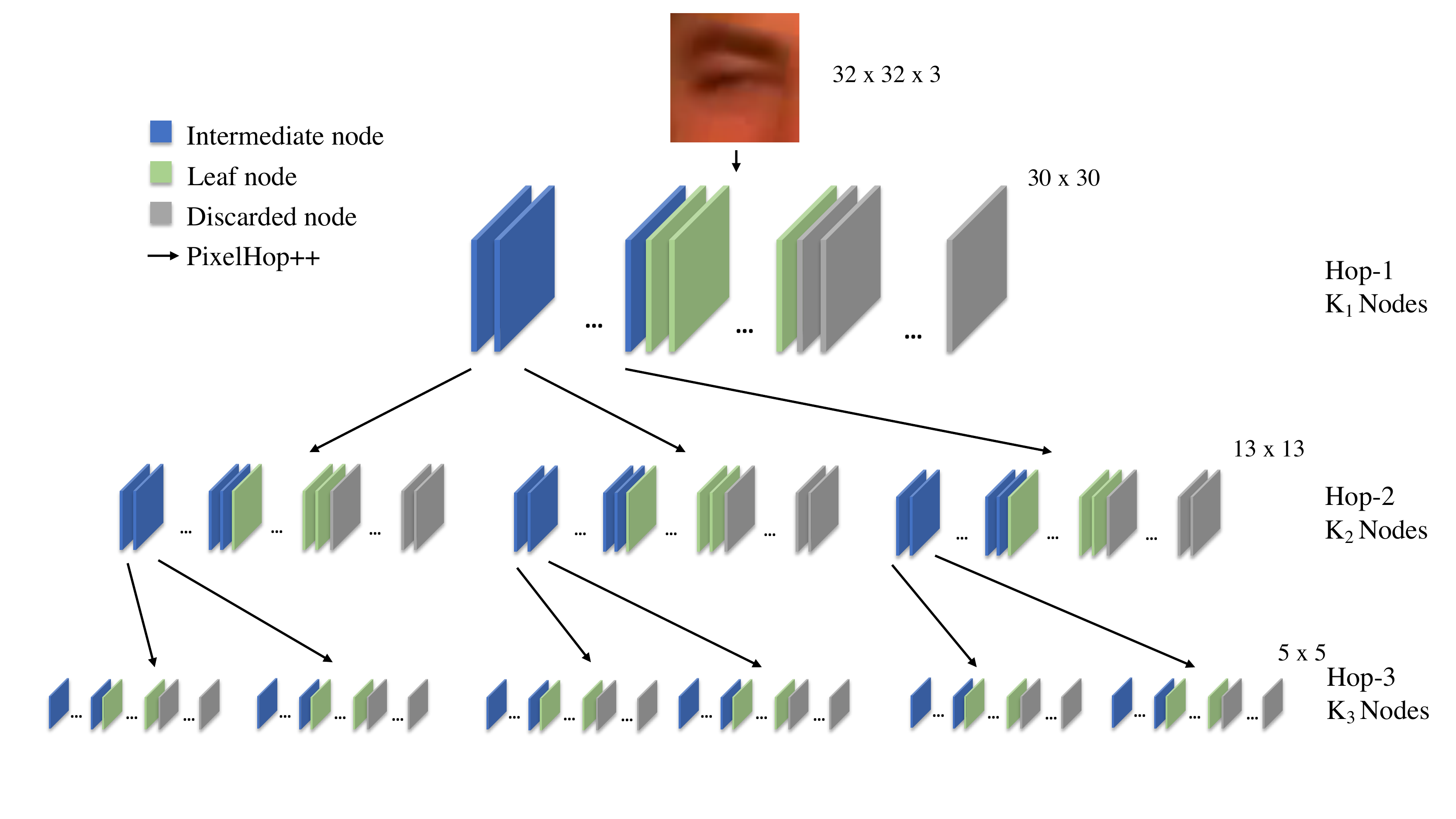}
\caption{Illustration of the c/w Saab transform.}\label{fig_tree}
\end{figure}

This scheme can be explained by the diagram in Fig.  \ref{fig_tree},
where a three-stage c/w Saab transform is illustrated. By following the
system in Fig. \ref{fig_ssl}, the root of the tree is the color image of
dimension $32 \times 32 \times 3$.  The local input vector to the first
hop has a dimension of $3\times3\times3=27$.  Thus, we can get a local mean
and 26 frequency components. We divide them into three groups:
low-frequency channels (in blue), mid-frequency channels (in green), and
high-frequency channels (in gray). Each channel can be represented as a
node in the tree.  Responses of high-frequency channels can
be discarded, responses of mid-frequency channels are kept, yet no
further transform is performed due to weak spatial correlations, and
responses of low-frequency channels will be fed into the next stage for
another c/w Saab transform due to stronger spatial correlations.
Responses in Hop-1, Hop-2 and Hop-3 are joint spatial-spectral
representations. The first few hops contain more spatial detail but have a
narrower view. As the hop goes deeper, it has less spatial detail but with a
broader view.  The channel-wise (c/w) Saab transform exploits channel
separability to reduce the model size of the Saab transform without
performance degradation.

\subsection{Feature Distillation Module}\label{subsubsec:cwSaab}

After feature extraction by PixelHop++, we derive a small and effective
set of features of a face. However, the output dimension of PixelHop++
is still not concise enough to be fed into a classifier. For example, the
output dimension of the first hop is 
$15 \times 15 \times K_{11}= 225 K_{11}$. We use two methods to further
distill the feature to get a compact description of a face being fake or real.

{\bf Spatial dimension reduction} 
Since input images are face patches from the same part of human faces, there exist
strong correlations between the spatial responses of $15\times 15$ for
a given channel. Thus, we apply PCA for further spatial dimension reduction. 
By keeping the top $N_{1} $ PCA components which contain about 90\% of the energy of the input images, we get a compact representation 
of $N_{1} \times K_{11}$ dimension.

{\bf Channel-wise Soft Classification}
After removing spatial and spectral redundancies, 
we obtain $K_{11}$ channels from each hop with a rather small spatial dimension $N_{1}$.
For each channel, we train a soft binary classifier to include the semantic 
meaning. The soft decision provides the probability of a specific channel being related to a fake video. 
Different classifiers could be used here in different situations.
In our model, the extreme gradient boosting classifier (XGBoost) is selected 
because it has a reasonable model size and is efficient to train to get a high AUC. The max-depth of XGBoost is set to one for all 
soft classifiers to prevent overfitting. 

For each face patch, we concatenate the probabilities of all channels  to 
get a description representing the patch. The output dimension becomes $K_{11}$, 
which is reduced by a large amount comparing to the output of PixelHop++.
It will be fed into a final classifier to determine the probability 
of being fake for this face patch.

\subsection{Ensemble Classification Module}

We integrate soft decisions from all facial regions (different face patches) and selected frames 
to offer the final decision whether a video clip is fake or real according to the following procedure.

{\bf Multi-Region Ensemble.} Since each facial region might have
different strength and weakness against different Deepfake manipulations
\cite{tolosana2020deepfakes}, we concatenate the probabilities of
different regions together.  In experiments
reported in Sec. \ref{sec:experiments}, we focus on three facial
regions. They are the left eye, the right eye, and the mouth. 

{\bf Multi-Frame Ensemble.} Besides concatenating different regions, 
for each frame, we concatenate the current frame and its adjacent 6 frames 
(namely, 3 frames before and 3 frames after) so as to incorporate
the temporal information.

Finally, for the whole video clip, we compute its probability of being 
fake by averaging the probabilities of all frames from the same video. Different approaches to aggregate frame-level probabilities can be used to determine the final decision.

\section{Experiments}\label{sec:experiments}

We evaluate the performance of DefakeHop on four Deepfake video
datasets: UADFV, FaceForensics++ (FF++), Celeb-DF v1, and Celeb-DF v2, and compare the results with state-of-the-art Deepfake detection methods.
UADFV does not specify the train/test split.  We randomly select 80\% for training and 20\% for testing.  FF++ and Celeb-DF provide the test set,
and all other videos are used for training the model.

{\bf Benchmarking Datasets.} Deepfake video datasets are categorized
into two generations based on the dataset size and Deepfake methods
used. The first generation includes UADFV and FF++. The second
generation includes Celeb-DF version 1 and version 2. Fake videos of the
second generation are more realistic, which makes their detection more
challenging.


\begin{table}[ht]
\caption{The AUC value for each facial region and the final ensemble result.} 
\label{tab:facial_region}
\begin{center}
\vspace{-0.3cm}
\begin{tabular}{ccccc}
\hline \hline
      & Left eye & Right eye & Mouth & Ensemble \\ \hline
UADFV & 100\% & 100\% & 100\% & 100\% \\ 
FF++ / DF  & 94.37\% & 93.73\% &  94.25\% & 97.45\% \\ 
Celeb-DF v1 & 89.69\% & 88.20\% & 92.66\% & 94.95\% \\ 
Celeb-DF v2 & 85.17\% & 86.41\% & 89.66\% & 90.56\%  \\ \hline
\end{tabular}
\vspace{-1cm}
\end{center}
\end{table}
\begin{table*}[ht]
\caption{Comparison of the detection performance of benchmarking methods
with the AUC value at the frame level as the evaluation metric. The
\textbf{boldface} and the underbar indicate the best and the second-best results, respectively. The \textit{italics} means it does not specify frame or video
level AUC.  The AUC results of DefakeHop is reported in both frame-level and
video-level. The AUC results of benchmarking methods
are taken from \cite{tolosana2020deepfakes_survey} and
\cite{li2020celeb}. 
$^{a}$ deep learning method, $^{b}$ non deep learning method. }\label{tab:compare}
\begin{center}
\begin{tabular}{c|c|cc|cc|c}
\hline\hline
&& \multicolumn{2}{c|}{1st Generation datasets} & \multicolumn{2}{c|}{2nd Generation datasets} & \multicolumn{1}{c}{}\\\hline
& Method & UADFV  &\makecell{FF++ / DF} &  \makecell{Celeb-DF \\ v1} & \makecell{Celeb-DF \\ v2}  & \makecell{Number of \\parameters} \\ \hline
Zhou {\em et al.}.(2017) \cite{zhou2017two}& InceptionV3$^{a}$  & 85.1\% & 70.1\%  & 55.7\% & 53.8\%& 24M \\
Afchar {\em et al.}.(2018) \cite{afchar2018mesonet} &Meso4$^{a}$ & 84.3\% & 84.7\%  & 53.6\% & 54.8\%& 27.9K \\
Li {\em et al.}.(2018) \cite{li2019exposing}  &FWA$^{a}$ (ResNet-50) & 97.4\% & 80.1  & 53.8\% & 56.9\%& 23.8M\\
Yang {\em et al.}.(2019) \cite{yang2019exposing} &HeadPose$^{b}$ (SVM)  & 89\% & 47.3\%  & 54.8\% & 54.6\%& - \\
Matern {\em et al.}.(2019) \cite{matern2019exploiting} & VA-MLP$^{b}$  & 70.2\% & 66.4\%  & 48.8\% & 55\%& - \\
Rossler {\em et al.}.(2019) \cite{rossler2019faceforensics++}  &Xception-raw$^{a}$ & 80.4\% & \textbf{99.7\%}  & 38.7\% & 48.2\%& 22.8M \\
Nguyen {\em et al.}.(2019) \cite{nguyen2019multi} &Multi-task$^{a}$  & 65.8\% & 76.3\%  & 36.5\% & 54.3\%& - \\
Nguyen {\em et al.}.(2019) \cite{nguyen2019use}  & CapsuleNet$^{a}$ & 61.3\% & 96.6\%  & - & 57.5\%& 3.9M \\
Sabir {\em et al.}.(2019) \cite{sabir2019recurrent} & \textit{DenseNet+RNN}$^{a}$  & - & \underbar{99.6\%}  & - & - & 25.6M \\
Li {\em et al.}.(2020) \cite{li2019exposing}& DSP-FWA$^{a}$ (SPPNet)  & \underbar{97.7\%} & 93\%  & - & 64.6\% & - \\
Tolosana {\em et al.}.(2020) \cite{tolosana2020deepfakes} & \textit{Xception$^{a}$}  & \textbf{100\%} & 99.4\%  & 83.6\% & - & 22.8M \\\hline
\multirow{2}{*}{Ours}  & DefakeHop (Frame) & \textbf{100\%} & 95.95\%  & \underbar{93.12\%} & \underbar{87.65\%}  & 42.8K\\
                       & DefakeHop (Video) & \textbf{100\%} & 97.45\%  & \textbf{94.95\%} & \textbf{90.56\%}  & 42.8K\\\hline
\end{tabular}
\end{center}
\end{table*}

\subsection{Detection Performance} 
\begin{figure}[htb]
\centering
\includegraphics[width=1\linewidth]{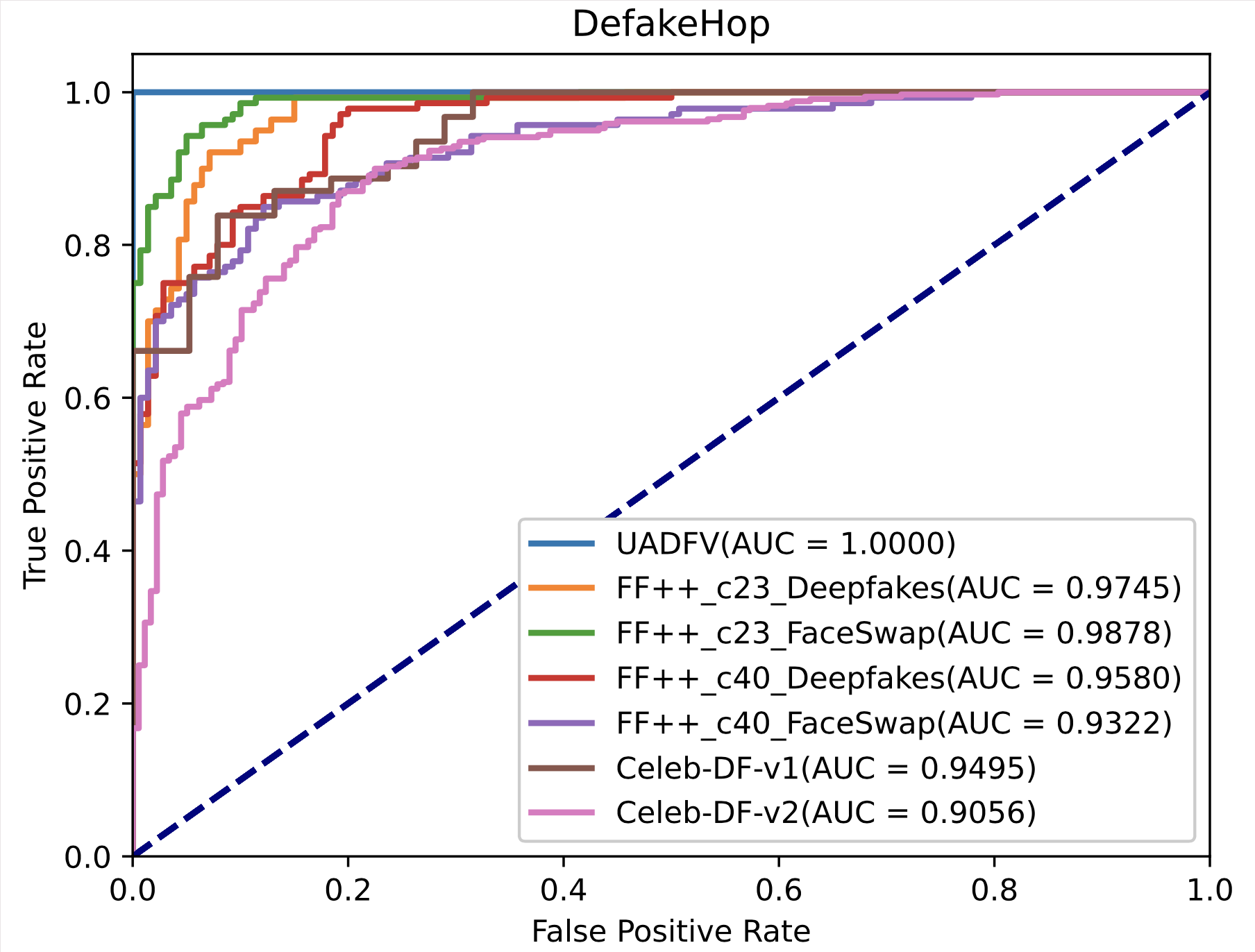}
\caption{The ROC curve of DefakeHop for different 
datasets.} \label{fig_roc}
\vspace{-0.5cm}
\end{figure}

DefakeHop is benchmarked with several other 
methods using the area under the ROC curve (AUC) metric in Table
\ref{tab:compare}. We report both frame-level and video-level AUC values
for DefakeHop. For the frame-level AUC, we follow the
evaluation done in \cite{li2020celeb} that considers key frames only
(rather than all frames). DefakeHop achieves the best performance on
UADFV, Celeb-DF v1, and Celeb-DF v2 among all methods. On FF++/DF, its
AUC is only 2.15\% lower than the best result ever reported.  DefakeHop outperforms other
non-DL methods \cite{yang2019exposing, matern2019exploiting} by a
significant margin.  It is also competitive against DL methods
\cite{tolosana2020deepfakes, nguyen2019use, nguyen2019multi,
rossler2019faceforensics++}. The ROC curve of DefakeHop with respect to
different datasets is shown in Fig.  \ref{fig_roc}. 

We compare the detection performance by considering different factors.

{\bf Facial Regions} The performance of DefakeHop with respect to
different facial regions and their ensemble results are shown in Table
\ref{tab:facial_region}.  The ensemble of multiple facial regions can
boost the AUC values by up to 5\%.  Each facial region has different strengths on various faces, and their ensemble gives the best result. 
 
{\bf Video Quality.} 
In the experiment, we focus on compressed videos since they
are challenging for Deepfake detection algorithms.
We evaluate the AUC performance of DefakeHop on videos with different qualities in Table \ref{tab:compression}. As shown in the table, the performance of
DefakeHop degrades by 5\% as video quality becomes worse. Thus,
DefakeHop can reasonably handle videos with different qualities.

\begin{table}[ht]
\caption{Comparison of Deepfake algorithms and qualities.}\label{tab:compression}
\begin{center}
\vspace{0.2cm}
\begin{tabular}{c|cc|cc}
\hline\hline
& \multicolumn{2}{c|}{FF++ with Deepfakes} & \multicolumn{2}{c}{FF++ with FaceSwap}\\\hline
& \makecell{HQ (c23)}  &\makecell{LQ (c40)} &  \makecell{HQ (c23)} & \makecell{LQ (c40)} \\ \hline
 Frame  & 95.95\% & 93.01\% & 97.87\% & 89.14\%  \\
 Video  & 97.45\% & 95.80\% & 98.78\% & 93.22\%  \\\hline
\end{tabular}
\vspace{-0.5cm}
\end{center}
\end{table}

\subsection{Model Size Computation} 
\quad {\bf PixelHop++ units}
Filters in all three hops have a size of 3x3. The 
maximum number of PixelHop++ units is limited to 10. For
Hop-1,  the input has three channels, leading to a 27D input vector.
For Hop-2 and Hop-3, since we use the channel-wise method, each input
has only 1 channel, leading to a 9D input vector. There are multiple 
c/w Saab transforms for Hop-2 and Hop-3. The channel would be selected
across multiple c/w Saab transforms. Therefore, it is possible to get 
more than 9 channels for Hop-2 and Hop-3.
At most $27 \times 10$, $9 \times 10$, $9 \times 10$ parameters are used 
in 3 PixelHop++ units.

{\bf Spatial PCA} Hop-1 is reduced from 225 to 45, Hop-2 is reduced from 49 to 30, and Hop-3 is reduced from 9 to 5. The number of parameters for spatial PCAs are $225 \times 45$, $49 \times 25$, and $9 \times 5$. The number of kept channels 45, 25, 5 is selected based on keeping about 90\% of energy.

{\bf XGBoost} The number of trees is set to 100. Each tree includes intermediate and leaf nodes. Intermediate nodes decide the dimension and boundary to split (i.e., 2 parameters per node) while the leaf nodes determine the predicted value (i.e., 1 parameter per node).
When the max-depths are 1 and 6, the number of parameters are 400 and 19,000. The max-depth of channel-wise and ensemble XGBoosts are set to 1 and 6, respectively. Thus, the total number of parameters for 30 channel-wise XGBoosts is 12,000 and the number of parameters of the ensemble XGBoost is 19,000.

\begin{table}[ht]
\vspace{-0.3cm}
\caption{The number of parameters for various parts.}\label{tab:parameters}
\begin{center}
\vspace{0.1cm}
\begin{tabular}{cccccc}

\hline
\hline
 Subsystem & \makecell{Number of Parameters}\\ \hline
Pixelhop++ Hop-1 & 270 \\ 
Pixelhop++ Hop-2 & 90  \\ 
Pixelhop++ Hop-3 & 90  \\ 
PCA Hop-1 & 10,125\\
PCA Hop-2 & 1,225 \\ 
PCA Hop-3 & 45 \\ 
Channel-Wise XGBoost(s) & 12,000 \\
Fianl XGBoost & 19,000 \\ \hline
\textbf{Total} & \textbf{42,845}\\ \hline
\end{tabular}

\end{center}
\vspace{-0.2cm}
\end{table}

The above calculations are summarized in Table \ref{tab:parameters}. The final model size, 42,845, is actually an upper-bound estimate since the
maximum depth in the XGBoost and the maximum channel number per hop are
upper bounded.

\begin{figure}
\centering
\includegraphics[width=1\linewidth]{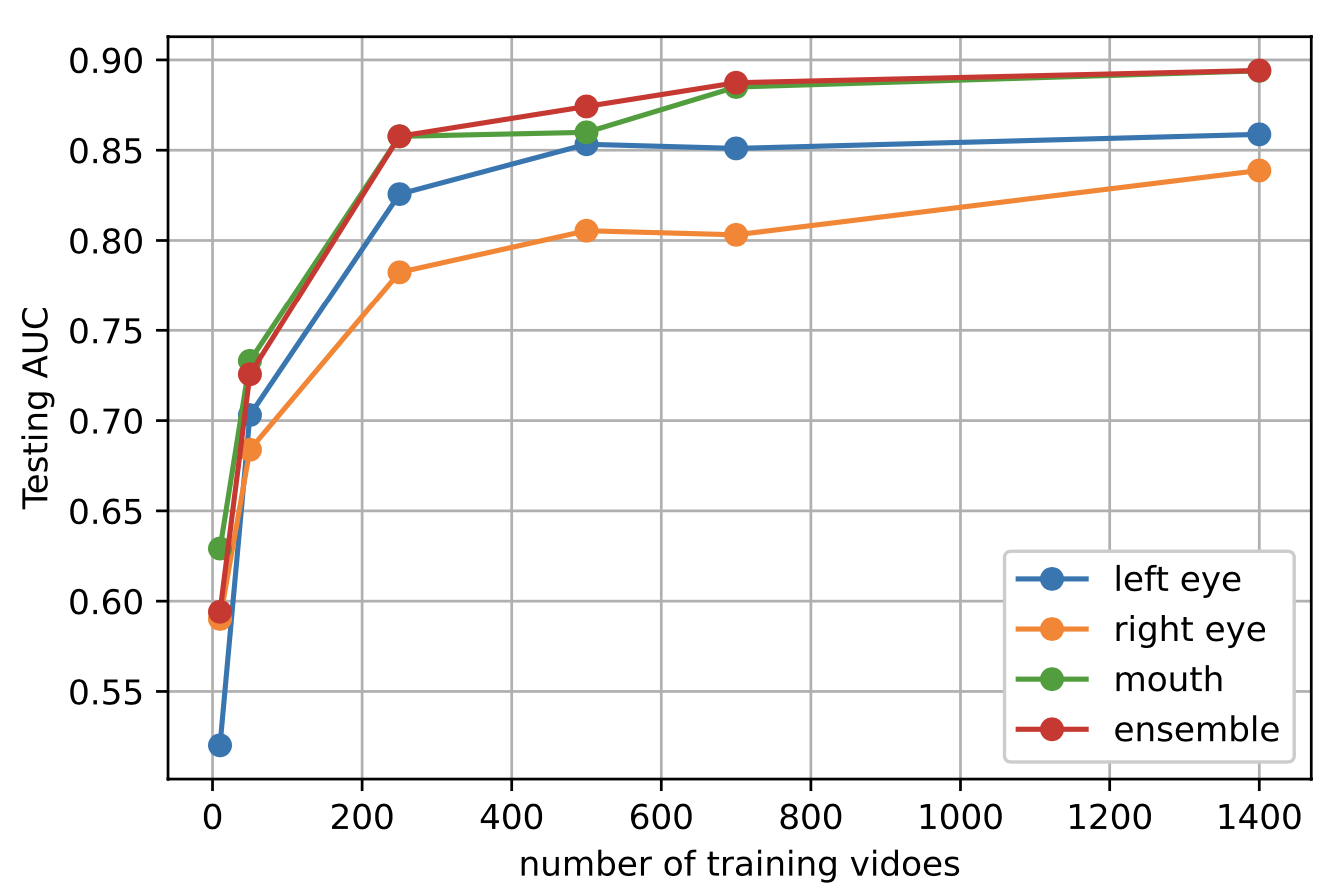}
\vspace{-0.5cm}
\caption{The plot of AUC values as a function of the training video number.}\label{fig_weak}
\vspace{-0.5cm}
\end{figure}

{\bf Train on a small data size} 
We observe that DefakeHop demands fewer training
videos. An example is given to illustrate this point. Celeb-DF v2 has
6011 training videos in total. We train DefakeHop with a randomly increasing
video number in three facial regions and see that DefakeHop can achieve
nearly 90\% AUC with only 1424 videos (23.6\% of the whole dataset). In
Fig. \ref{fig_weak} the AUC of test videos is plotted as a
function of the number of training videos of Celeb-DF v2.  As shown in the
figure, DefakeHop can achieve about 85\% AUC with less than 5\% (250
videos) of the whole training data. 

\section{Conclusion}\label{sec:conclusion}

A light-weight high-performance method for Deepfake detection, 
called DefakeHop, was proposed.
It has several advantages: a smaller model size, 
fast training procedure, high detection AUC and 
needs fewer training samples. Extensive experiments were conducted 
to demonstrate its high detection performance. 

\bibliographystyle{IEEEbib}
\bibliography{refs}

\end{document}